# An Approach of Improving Student's Academic Performance by using K-means clustering algorithm and Decision tree

Md. Hedayetul Islam Shovon
Department of Computer Science and Engineering
Rajshahi University of Engineering & Technology
Rajshahi-6204, Bangladesh

Mahfuza Haque
Department of Computer Science and Engineering
Rajshahi University of Engineering & Technology
Rajshahi-6204, Bangladesh

*Abstract*—Improving student's academic performance is not an easy task for the academic community of higher learning. The academic performance of engineering and science students during their first year at university is a turning point in their educational path and usually encroaches on their General Point Average (GPA) in a decisive manner. The students evaluation factors like class quizzes mid and final exam assignment lab - work are studied. It is recommended that all these correlated information should be conveyed to the class teacher before the conduction of final exam. This study will help the teachers to reduce the drop out ratio to a significant level and improve the performance of students. In this paper, we present a hybrid procedure based on Decision Tree of Data mining method and Data Clustering that enables academicians to predict student's GPA and based on that instructor can take necessary step to improve student academic performance

## I. INTRODUCTION

Graded Point Average (GPA) is a commonly used indicator of academic performance. Many universities set a minimum GPA that should be maintained. Therefore, GPA still remains the most common factor used by the academic planners to evaluate progression in an academic environment. Many factors could act as barriers to student attaining and maintaining a high GPA that reflects their overall academic performance, during their tenure in university. These factors could be targeted by the faculty members in developing strategies to improve student learning and improve their academic performance by way of monitoring the progression of their performance [1]. With the help of clustering algorithm and decision tree of data mining technique it is possible to discover the key characteristics for future prediction. Data clustering is a process of extracting previously unknown, valid, positional useful and hidden patterns from large data sets. The amount of data stored in educational databases is increasing rapidly. Clustering technique is most widely used technique for future prediction. The main goal of clustering is to partition students into homogeneous groups according to their characteristics and abilities (Kifaya, 2009). These applications can help both instructor and student to enhance the education quality. This study makes use of cluster analysis to segment students into groups according to their characteristics [2]. Decision tree analysis is a popular data mining technique that can be used to explain different variables like attendance ratio and grade ratio. Clustering is one of the basic techniques often used in analyzing data sets [3]. This study makes use of cluster analysis to segment students in to groups according to their characteristics and use decision tree for making meaningful decision for the student's.

## 2. Methodology

A. *Data Clustering*
Data Clustering is unsupervised and statistical data analysis technique. It is used to classify the same data into a homogeneous group. It is used to operate on a large data-set to discover hidden pattern and relationship helps to make decision quickly and efficiently. In a word, Cluster analysis is used to segment a large set of data into subsets called clusters. Each cluster is a collection of data objects that are similar to one another are placed within the same cluster but are dissimilar to objects in other clusters.

a. *Implementation Of K-Means Clustering Algorithm*
K-Means is one of the simplest unsupervised learning algorithms used for clustering. K-means partitions "n" observations in to k clusters in which each observation belongs to the cluster with the nearest mean. This algorithm aims at minimizing an objective function, in this case a squared error function. The algorithm and flow-chart of K-means clustering is given below…

**Algorithm 1** Basic K-means Algorithm.
1: Select $K$ points as the initial centroids.
2: **repeat**
3:    Form $K$ clusters by assigning all points to the closest centroid.
4:    Recompute the centroid of each cluster.
5: **until** The centroids don't change

Fig.1 Traditional K-Means Algorithm [4].

From the algorithm it is easily seen that, initially we have only the raw data. So, it is clustered around a single point. If the cluster number K is fixed then we need to cluster around that point. If the cluster is not fixed then it is continued until the centered is not changed. Initially the students are all in a same group. But when K-means clustering is applied on it then it clusters the student's into three major categories, one is good, one is medium, and the other is low standard student.

The flow chart of the k-means algorithm that means how the k-means work out is given below.

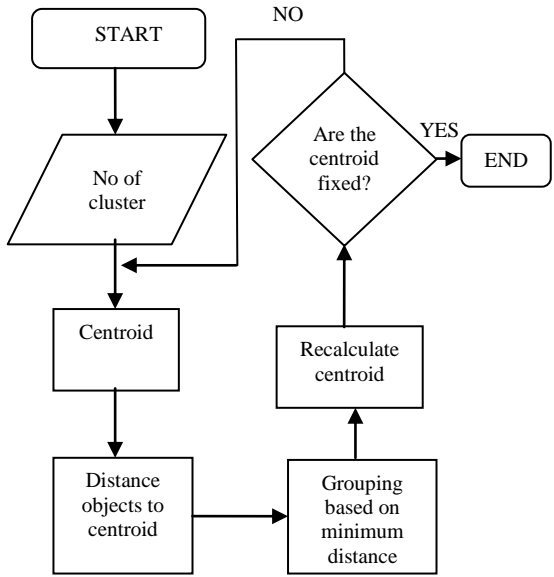

Fig.2 Flow-Chart Of K-Means Clustering.

### B. Data mining

Data mining, also popularly known as Knowledge Discovery in Database, refers to extracting or "mining" knowledge from large amounts of data. Data mining techniques are used to operate on large volumes of data to discover hidden patterns and relationships helpful in decision making. The sequences of steps identified in extracting knowledge from data are shown in fig.3

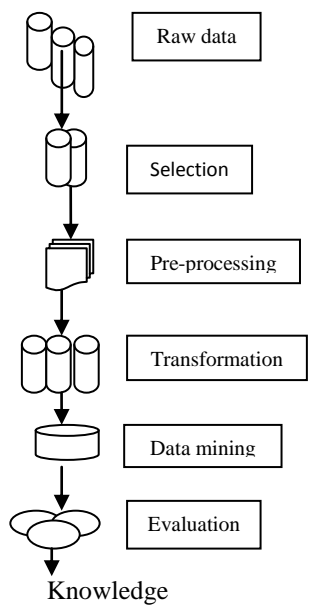

Fig.3 Steps Of Knowledge Extraction

*I. Decision Tree*

Decision tree induction can be integrated with data warehousing techniques for data mining. A decision tree is a predictive node ling technique used in classification, clustering, and prediction tasks. A decision tree is a tree where the root and each internal node are labeled with a question. The arcs emanating from each node represent each possible answer to the associated question. Each leaf node represents a prediction of a solution to the problem under consideration.

The basic algorithm for decision tree induction is a greedy algorithm that constructs decision trees in a top-down recursive divide-and-conquer manner. Decision Tree Algorithm: generate a decision tree from the given training data.

1. Create a node N
2. If samples are all of the same class, C then
3. Return N as a leaf node labeled with the class C;
4. If attribute-list is empty then
5. Return N as a leaf node labeled with the most common class in samples.
6. Select test-attribute, the attribute among attribute-list with the highest information gain;
7. Label node N with test-attribute;
8. For each known value $a_i$ of test-attribute.
9. Grow a branch from node N for the condition test attribute = $a_i$ ;
10. Let Si be the set of samples for which test-attribute = $a_i$;
11. If Si is empty then
12. Attach a leaf labeled with the most common class in samples;
13. Else attach the node returned by generate-decision-tree ($S_i$,attribute-list-attribute);

Each internal node tests an attribute, each branch corresponds to attribute value, and each leaf node assigns a classification.

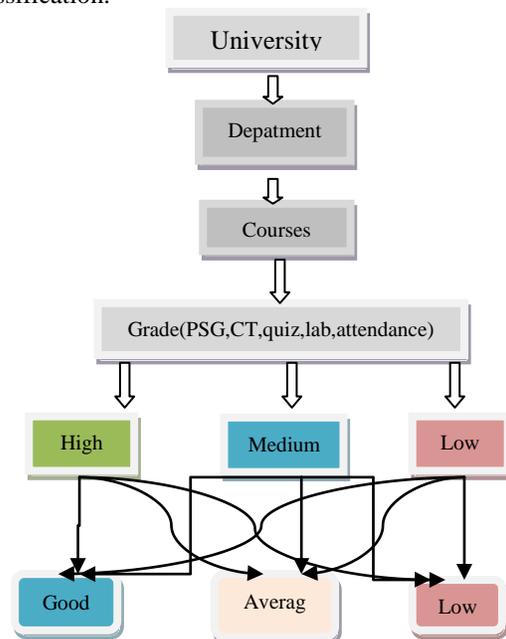

Fig.4 Decision Tree.

*(IJACSA) International Journal of Advanced Computer Science and Applications,*
*Vol.3, No. 8, 2012*

From 50 training sample here only 20 samples are shown.

Table 1: Training Sample.

| Roll | GPA | CT | Attendance | Assignment | Lab_per | Quiz |
|---|---|---|---|---|---|---|
| 1 | 3.89 | 19 | 10 | Y | good | Y |
| 2 | 3.53 | 12 | 10 | Y | avg | Y |
| 3 | 3.2 | 10 | 10 | Y | good | Y |
| 4 | 3.6 | 16 | 10 | N | avg | N |
| 5 | 3.54 | 12 | 10 | Y | bad | Y |
| 6 | 3.5 | 10 | 10 | Y | good | N |
| 7 | 3 | 5 | 5 | N | avg | Y |
| 8 | 3.74 | 12 | 10 | Y | good | N |
| 9 | 3.67 | 14 | 10 | Y | good | Y |
| 10 | 2.05 | 2 | 6 | N | bad | N |
| 11 | 3.25 | 3 | 6 | N | avg | Y |
| 12 | 3.56 | 5 | 8 | N | good | Y |
| 13 | 3.2 | 9 | 5 | Y | avg | Y |
| 14 | 3.5 | 14 | 10 | Y | avg | Y |
| 15 | 3.2 | 10 | 10 | N | good | Y |
| 16 | 2.99 | 0 | 0 | Y | avg | N |
| 17 | 2.98 | 0 | 0 | N | avg | N |
| 18 | 3.87 | 3 | 5 | Y | avg | N |
| 19 | 3.45 | 8 | 8 | Y | avg | N |
| 20 | 3.21 | 9 | 6 | N | avg | N |

## II. RESULT AND DISCUSSION

From the training data GPA and the attendance ration of the student is given below

**Graph.1:** Shows the relationship between GPA and Attendance ratio.

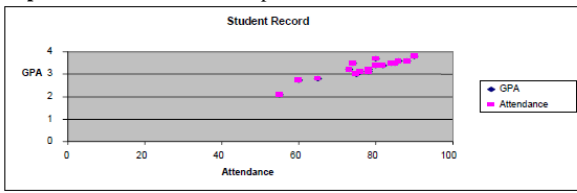

If we apply K-means clustering algorithm on the training data then we can group the students in three classes "High" "Medium" and "Low" according to their new grade. New grade is calculated from the previous semester grade that means external assessment and internal assessment. The table and corresponding graph is given below.

Table 2. Percentage of students according to GPA.

| Class | GPA | No of student | Percentage |
|---|---|---|---|
| 1 | 2.00-2.20 | 5 | 8.33 |
| 2 | 2.20-3.00 | 10 | 16.67 |
| 3 | 3.00-3.32 | 17 | 28.33 |
| 4 | 3.32-3.56 | 15 | 25 |
| 5 | 3.56-4.00 | 13 | 21.67 |

Here, I cluster student among their GPA that means, from GPA 2.00- 2.20 we have 8.33% student. From 2.20-3.00 student percentage is 16.67%.

From 3.00-3.32 we have 28.33%. From 3.32-3.56 percentage is 25% .The percentage is 21.67% between GPA 3.56-4.00.

The graphical representation of GPA and the percentage of student's among the student are given below.

**Graph 2:** Number and percentage of students regarding to GPA

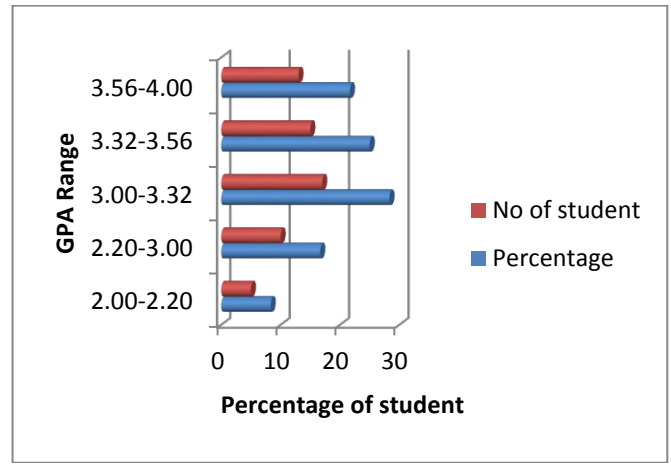

Table 2

| Class | GPA | No of student | Percentage |
|---|---|---|---|
| High | >=3.50 | 28 | 46.67 |
| Medium | 2.20<=GPA <3.5 | 27 | 45 |
| Low | <=2.20 | 5 | 8.33 |

After clustering the student, we group the student into three categories. One is High, second is Medium, and the last one is Low.

Graphical representation of these three categories is given below.

**Graph 3:** Shows the percentage of students getting high, medium and low GPA

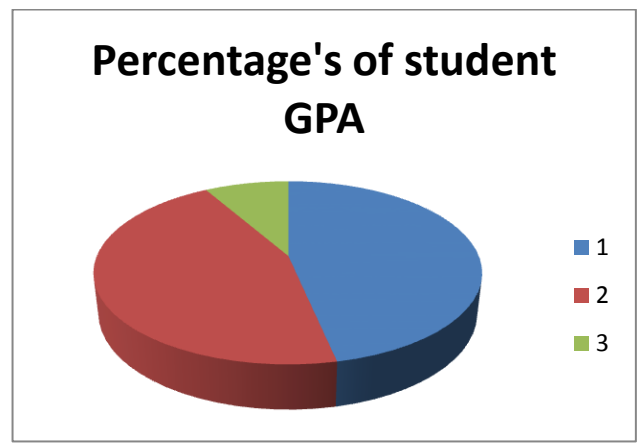

If we apply data mining technique decision tree then it will help us to make correct decision about the student which is need to take by the instructor. The decision step is given below.

Table 3: Decision based on the student categories:

| Step No | Grade | Effort |
|---------|-------|--------|
| S-01 | A+ | He/She is a good student. Need not to take special care. |
| S-02 | A,A- | Is not so good. Need to take care of CT & Quiz. |
| S-03 | B+,B | Is a medium student. Should take care of CT,quiz and lab performance also. |
| S-04 | Below B grade | Is a lower standard student. Need lot of practice of his/her lesson and also take care of all the courses ct,lab,quiz ,attendance carefully. |

### III. CONCLUSION AND FUTURE WORK

In this study we make use of data mining process in student's database using k-means clustering algorithm and decision tree technique to predict student's learning activities. We hope that the information generated after the implementation of data mining and data clustering technique may be helpful for instructor as well as for students. This work may improve student's performance; reduce failing ratio by taking appropriate steps at right time to improve the quality of education. For future work, we hope to refine our technique in order to get more valuable and accurate outputs, useful for instructors to improve the students learning outcomes.

AUTHORS PROFILE

**Md. Hedayetul Islam Shovon** was born in 1986 in Bangladesh. He received his B.Sc Engineering degree from Rajshahi University of Engineering and Technology (RUET) in 2009. He joined in the department of Computer Science and Engineering of RUET in October 2009. Since then he involved in different research oriented activities. His research interests include Image Processing, Data Clustering, Computer Vision and Optimization etc.

**Mahfuza Haque** was born in Bangladesh. He received his B.Sc Engineering degree from the department of Computer Science and Engineering of Rajshahi University of Engineering and Technology (RUET) in 2012. His research interest includes Pattern recognition, Data clustering, Data mining.